\documentclass[11pt]{article}

\usepackage{acl}

\usepackage{times}
\usepackage{latexsym}

\usepackage[T1]{fontenc}
\usepackage[utf8]{inputenc}

\usepackage{microtype}

\usepackage{inconsolata}
\usepackage{listings}
\usepackage{float}
\usepackage{graphicx}
\usepackage{parskip}
\usepackage{indentfirst}
\usepackage{tabularx}
\usepackage{url}
\usepackage{booktabs}
\usepackage{mdframed}
\usepackage[frozencache,cachedir=.]{minted}
\usepackage[most]{tcolorbox} 
\usepackage{multicol}

\usepackage{enumitem}

\title{Efficient Medical Question Answering with Knowledge-Augmented Question Generation}

\author{Julien Khlaut $^{*1}$, \quad  Corentin Dancette$^{*1}$, \quad Elodie Ferreres$^{*1}$, \\ \bf{Alaedine Bennani$^{*2}$,\quad Paul Hérent$^{1}$, \quad Pierre Manceron$^{1}$} \\
  $^{1}$Raidium \\ $^2$ Service de médecine vasculaire,  Hôpital européen Georges Pompidou (HEGP), \\ AP-HP, Université Paris-Cité, Paris, France \\
  $^{1}$\texttt{first.last@raidium.fr} \hspace{2mm} $^{2}$\texttt{alaedine.benani@aphp.fr}
  }

\begin{document}
\maketitle

\def\thefootnote{*}\footnotetext{Equal Contribution}

\def\thefootnote{\arabic{footnote}}

% text text text\footnote{normal footnote}

\begin{abstract}
In the expanding field of language model applications, medical knowledge representation remains a significant challenge due to the specialized nature of the domain. Large language models, such as GPT-4 \cite{openaiGPT4TechnicalReport2023}, obtain reasonable scores on medical question answering tasks, but smaller models are far behind.
In this work, we introduce a method to improve the proficiency of a small language model in the medical domain by employing a two-fold approach. We first fine-tune the model on a corpus of medical textbooks. Then, we use GPT-4 to generate questions similar to the downstream task, prompted with textbook knowledge, and use them to fine-tune the model. Additionally, we introduce ECN-QA, a novel medical question answering dataset containing ``progressive questions'' composed of related sequential questions. We show the benefits of our training strategy on this dataset.
The study's findings highlight the potential of small language models in the medical domain when appropriately fine-tuned.
The code and weights are available at \url{https://github.com/raidium-med/MQG}.
\end{abstract}

\section{Introduction}
Deep Learning led to a breakthrough in natural language processing, reaching human performances on many tasks like question answering or translation. However, their performances are still subpar in complex domains, such as medicine. This domain presents unique challenges, mainly due to its specialized vocabulary, complex concepts, and fast-changing medical literature. Language-based medical tasks, such as medical question answering, require vast knowledge and reasoning abilities to make correct diagnoses.
Traditional language models (LMs), while effective in general language processing, struggle when faced with medical knowledge learning mainly because sufficient data for medical knowledge is not necessarily readily available for training. 
Moreover, in the context of language models, their number of parameters often plays a pivotal role in performances. Large models, although powerful, come with high computational costs and resource requirements, both for training and inference, making them less accessible and practical for widespread use. On the other hand, small models, which are more economical, face challenges in generalization and adapting to specialized domains like medicine. These models require careful fine-tuning to grasp the depth and breadth of medical knowledge effectively.
The diversity of general, non-medical datasets on which LMs are trained poses another challenge. These datasets, encompassing a wide array of topics and styles, do not specifically cater to the medical domain. As a result, small models trained on such datasets might fail to develop the necessary understanding for answering more specialized medical questions.

Therefore, we tackle these issues for medical question answering tasks. First, we design a new dataset, ECN-QA. Existing medical question answering (QA) datasets such as MedQA~\cite{medQA2020} and others \cite{PubMedQADatasetBiomedical2019}, \cite{MedMCQALargescaleMultiSubject2022} are usually single-question multiple answers, which do not encompass the complexity of making a medical diagnosis, which requires multiple turns of questions. 
Our dataset is based on the French medical residency examination and contains multiple related questions that require models to remember previous questions and reasoning over multiple steps.
We then propose a method to train small to mid-size language models for medical question answering. We leverage a corpus of medical textbooks for pre-training. The pre-training set is enriched with specialized questions generated by large language models prompted with medical data from books. This helps to specialize the model on the target task with a small amount of original data. Our code will be made available online.

\section{Datasets}

\subsection{ECN-QA Dataset}
We design ECN-QA, a medical question answering dataset.
The questions are collected from FreeCN\footnote{\url{https://www.freecn.io}}, a website established by French medical students to facilitate ECN (\textit{Examen Classant National}, the national ranking exam before medical residency), with their authorization.
This website includes questions from past exams and additional questions (``custom'' questions) to simulate exam conditions and aid in studying.

The ECN exams themselves consist of two parts. The first part, known as \textit{Individual Questions} (IQ), features general medicine questions with 5 possible answers. Among these answers, one or multiple may be correct, and candidates must identify the true ones. We display an example in Table ~\ref{tab:example-qi}.
The dataset contains 4481 IQ, 721 of which come from the historical data of previous exams. The rest, the ``custom'' subset, contains 3760 additional IQ-like questions created by the FreeCN team to help students prepare for the exam.
The second part is known as \textit{Progressive Questions} (PQ), which features clinical cases. Each PQ consists of an introduction followed by a series of successive questions. Similar to the IQ section, these questions also offer 5 possible answers, with 1 to 5 correct answers. A single PQ can contain numerous successive sub-questions, sometimes more than 20. We have 1050 sub-questions in all PQ. We show an example in Table \ref{tab:example-dp} of Appendix~\ref{sec:app:example-case}. We also show a whole progressive question in Appendix~\ref{sec:app:full-dp}. We use the accuracy as our evaluation metric. Each proposition in the question is answered separately and gets a score of 0 or 1.
The accuracy is then averaged over the five propositions, i.e., for one question, the possible score can be 0, 0.2, 0.4, 0.6, 0.8, or 1.0. For example, in Table \ref{tab:example-qi}, if the model answers a, b, c, e as wrong and d as right, it would have one error since c is right. The accuracy would, therefore, be 0.8. If the model answers a and e as wrong and b, c, and d as right, it would also have an accuracy of 0.8.

\setlist[itemize]{leftmargin=5mm}
\begin{table}[h]
\footnotesize
\centering
\begin{tabular}
{|m{.90\linewidth}|}
\hline
    \vspace{1mm}
     \textbf{Question}: A woman of Martinican origin has just given birth. The child's father is also of Martinican origin. The child has a cleft lip and palate. With regard to regulatory newborn screening of this child, what is the exact proposal(s)?
    \textbf{Propositions}: 
    \begin{enumerate}[label=(\alph*)]
        \item Phenylketonuria is the only disease of amino acid and organic acid metabolism currently being screened for newborn in France
        \item General screening test can detect hypothyroidism of pituitary origin
        \item \textbf{This couple can refuse the screening after information}
        \item \textbf{Completion before 48 h of life decreases the sensitivity and/or specificity of the screening test}
        \item Targeted screening for sickle cell disease is not indicated in this child
         \vspace{-3mm}
\end{enumerate} \\
\hline
\end{tabular}
\caption{Example of Individual Question (IQ) in the ECN, translated to English. Correct answers are in bold.}
\label{tab:example-qi}
\end{table}

All the original data is in French, but all models are pre-trained using mostly English data. Therefore, we translate all the questions and answers into English using the Azure AI Translation API\footnote{\url{https://learn.microsoft.com/en-us/azure/ai-services/translator/}}.

\subsection{Medical Textbooks}
Additionally, we use classical French medical textbooks designed for medical students, containing comprehensive medical knowledge and established protocols for managing various medical conditions. We detail in Section~\ref{sec:app:pdf} how we extract sections from medical textbooks in PDF format. 

In total, we worked with 17,509 PDF files. We grouped text in sections rather than pages, recognizing that a single topic might span multiple pages and should not be truncated. The sections are defined by the book titles and correspond to chapters or important parts. This approach resulted in a total of 234,495 sections. The full dataset is composed of 174,242,531 tokens (with the GPT-3 tokenizer). We detail how we extract sections from PDF files in Appendix~\ref{sec:app:pdf}.

We use them for pre-training, and to generate additional questions, as explained in Section~\ref{sec:method}.

% \subsection{Evaluation Metric}
% We evaluate the 
% The correction process for these questions is based on a system of ``discrepancy.'' Specifically, candidates earn 1 point for providing answers with no discrepancies, 0.5 points for one discrepancy, 0.2 points for two discrepancies, and no points for three discrepancies or more in their responses.

\section{Method}
\label{sec:method}

We detail our training strategy in this section. The strategy is depicted in Figure~\ref{fig:main}. We detail related works in the Appendix ~\ref{sec:app:related}.

\subsection{Baseline Model}
For our baseline, we use the BioMedLM model~\cite{StanfordCRFM}. This 2.7-billion-parameter model is built upon the GPT-2 architecture \cite{radfordLanguageModelsAre} and has been trained on a substantial corpus of medical and biological data. BioMedLM's specialized biomedical tokenizer sets it apart, enhancing its comprehension of specialized terminology. 
BioMedLM's training data contains all PubMed abstracts and full documents from The Pile \cite{pile}, ensuring a rich knowledge base. 
Notably, BioMedLM reported state-of-the-art scores on the MedQA \cite{medQA2020} dataset.

However, this model does not possess the scale needed to achieve impressive zero-shot generalization on new tasks, and medical question answering datasets are limited in scale. Therefore, we aim to train it on specific high-quality data that resembles our benchmark. As our training dataset is small (4967 questions), we propose a method to augment it with question generation using a large language model prompted by some medical knowledge extracted from textbooks.

\begin{figure}[h]
    \centering
    \includegraphics[width=\linewidth]{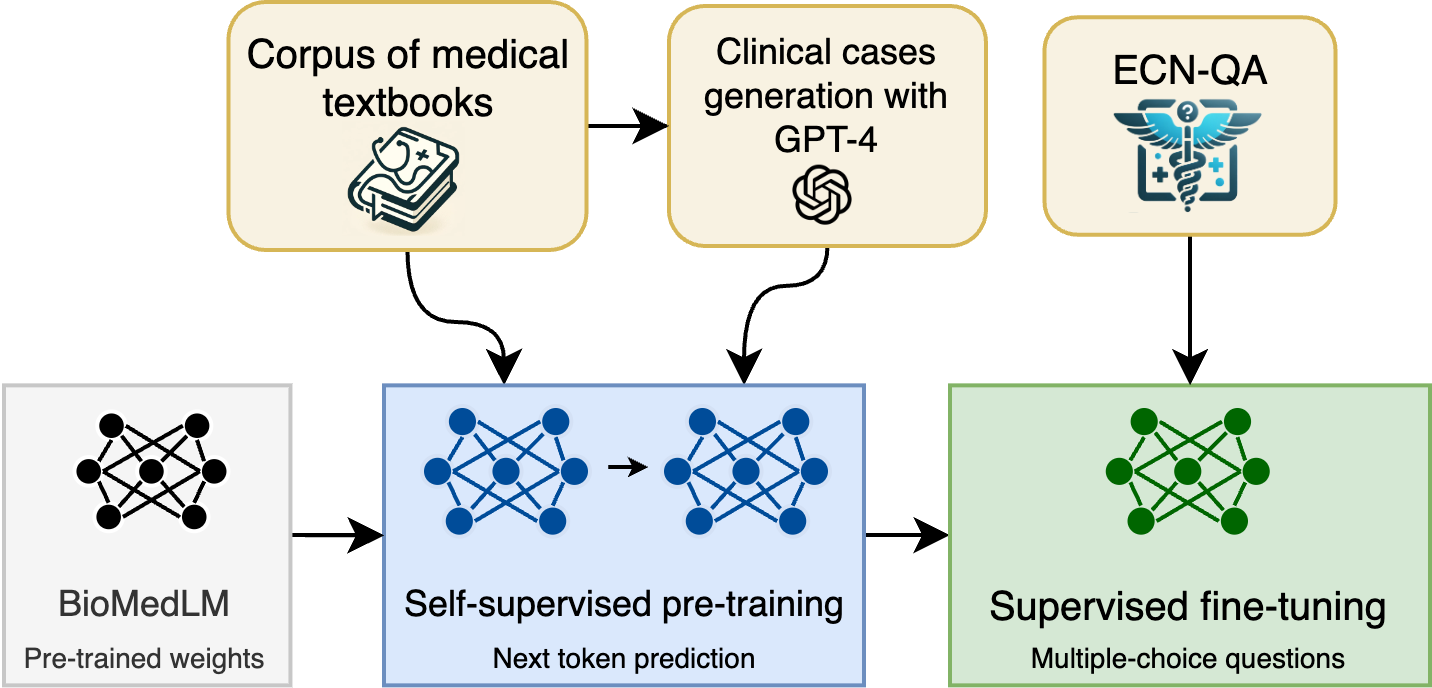}
    \caption{Our training strategy. Starting from an existing language model such as BioMedLM, we continue the pre-training on our corpus of medical textbooks. Then, we use GPT-4, prompted with knowledge from the textbooks, to generate clinical cases that are used to fine-tune the model.}
    \label{fig:main}
\end{figure}

\subsection{Questions Generation}

Our objective is to create cases that closely resemble genuine ECN cases, as this offers the most effective training for the model. 
The format we desire for these cases closely resembles that of the progressive questions: an introduction, a list of questions and their possible answers, with a label (true or false) for each answer.

To create our clinical cases, we concatenate several prompts using different approaches. The design of each prompt begins with adopting the prompt used by FreeCN, which primarily comprises an introduction to the task. We refer to this as the pre-prompt. Next, we compile a list of all the specific details we want the case to encompass. This list is informed by the insights of medical experts and the manner in which they typically structure questions for the ECN. We refer to this list as the “constitution.”
When we initially applied this approach, we encountered somewhat disappointing results. The clinical cases exhibited two major shortcomings. First, they often had very similar subjects, causing the model to struggle to generate diverse cases. Additionally, the main issue was that the questions posed were consistently identical, revolving around topics such as “What is the diagnosis?” or “What tests would you conduct for confirmation?” and “How would you manage the patient?”. To solve this issue and to introduce diversity in disease scenarios, we also supplied it with specific knowledge that could be utilized to construct these cases. This section was named the “knowledge part,” and it drew upon information extracted from sections of medical books. An example can be seen in Annex~\ref{tab:final-prompt}.

We introduce an additional ``justification'' field. This component explains why a particular answer to a question is deemed suitable or not.

We build our pipeline using the OpenAI API, employing the GPT-4 model~\cite{Achiam2023GPT4TR} to generate clinical cases.
We use the GPT-4 function calling JSON mode, which allows us to specify the output structure. 

Following this approach, we generated a dataset with GPT-4, containing about 10,237,240 tokens. In some instances, the dataset underwent meticulous filtering procedures to rectify issues such as missing or alternative fields.
This approach is inspired by phi~\cite{gunasekarTextbooksAreAll2023,liTextbooksAreAll2023} and Orca~\cite{mitraOrcaTeachingSmall2023,mukherjeeOrcaProgressiveLearning2023}.

We gathered feedback and validation from the FreeCN team, composed of medical doctor students, for assistance and insights to ensure the quality of the questions. The prompt given to GPT-4 is displayed in Appendix~\ref{sec:app:prompt}, and an example of a generated progressive question in Appendix~\ref{sec:app:generated-dp}.

\subsection{Pre-training}

The initial phase involves pre-training the model on a dataset, partly composed of medical books and the additional generated questions. We start from BioMedLM's weights and use a next-token prediction loss to pre-train for three epochs.
After training on the books, the model is further trained on the generated cases. The 160,889 generated questions are composed of 10,237,240 tokens. The training is performed on one case at a time and the final loss is computed only on the model's answer and justification. Since the context length of BioMedLM is 2048, we truncate more prolonged cases. The training parameters are detailed in Appendix~\ref{sec:appendix:implementation}.

\subsection{Fine-tuning}

Following the pre-training phase, the next step is fine-tuning the model on the ECN-QA dataset. For fine-tuning, the dataset is split into 90 \% for training and 10\% for testing set.
There are multiple ways of getting the model to output an answer, for example, generating tokens with a specific format. However, since generating consistent word-by-word answers proved challenging for the model, often resulting in gibberish rather than accurate responses, we opted for a more traditional approach during fine-tuning.
Similarly to previous work~\cite{StanfordCRFM}, a classification head was added to the model. It operates at the proposition level: the model takes as input the question and a single proposition among the five. It then has to predict if the proposition is right or wrong, as a binary classification task.
% There are two options here: predicting directly the right propositions, among the 5 proposed, or predicting, for each proposition, if it is right or wrong. operate in different ways for instance predicting for each question the right and wrong proposition. But during our testing, we found that operating at the proposition level gave better results. So our model will see each proposition independently and determine if a proposition is true or false. This means that it can not use its answer to other propositions. But there still are multiple approaches to get answers from that head.}
One possible approach to this binary classification involves predicting a single scalar value for each answer, training it with binary cross-entropy, and selecting a threshold value for inference.
Another approach consists of adding the words ``true" or ``false" to the end of the sentence, feeding both sentences to the model, and selecting the answer with the highest score. Empirically, the second approach provided the best results. This modification allowed us to obtain more reliable responses from the model during evaluation.

\section{Results}

\subsection{Evaluation of GPT models}

We first evaluate the GPT models on our dataset to obtain baseline scores. For both GPT-3.5 and GPT-4 models, the 2023-12-01 version of the API is used (available on Azure).

We encountered occasional issues during evaluation, where specific prompts may have been blocked, possibly due to sensitive subjects like pediatric medicine. In such cases, we considered the model's response incorrect. The prompts were designed to be straightforward, typically asking the model to provide a true or false answer. Moreover, questions were asked in English using the translated dataset.

\begin{table}[h]
\centering
\footnotesize
\begin{tabular}{lc}
\toprule
 Model    & Accuracy \\ % & Discordance  \\
\midrule
GPT-3.5   & 69.36  \\ % & 0.42 \\
\midrule
GPT-4     & 79.04  \\ % & 0.58  \\
\midrule
GPT-4-32k & 78.97 \\ % & 0.57  \\ 
\midrule
GPT-4-32k 5 few shot & 81.42  \\ %  & 0.62   \\   
\bottomrule
\end{tabular}
\caption{Results on the all evaluation dataset}
\label{tab:gpt-results}
\end{table}

The results are presented in Table~\ref{tab:gpt-results}.
GPT-4's performances on our dataset are similar to those on MedQA and USMLE, reaching zero-shot performances of around 74\% \cite{noriCapabilitiesGPT4Medical2023a}. Overall, GPT-4 and its 32k-context variant is the strongest model. Additionally, we confirm \cite{nori2023capabilities}'s findings that adding some questions in the prompt (\textit{few shot}) increases the accuracy, in our case, by around 2.5 points. 

% \begin{figure}[H]
%     \centering
%     \subfloat[\centering Loss obtained in pre-training]{{\includegraphics[width=13cm]{images/loss_cases.png} }}%
%     \qquad
%     \subfloat[\centering Eval Loss obtained in pre-training]{{\includegraphics[width=13cm]{images/eval_cases.png} }}%
%     \caption{Results in pre-training using normal SSL on the generated cases with or without the books}%
% \end{figure}

\subsection{Main Results}
\begin{table}[h] %[H]
\centering
\begin{tabular}{lc}
\toprule
Model & Accuracy \\% Discordance \\
\midrule
BioMedLM  & 67.74   \\% & 0.3827 \\
\midrule
BioMedLM + Books & 69.65 \\%    & 0.4160    \\
\midrule
BioMedLM + MQG & 68.62    \\% & 0.3963     \\ 
\midrule
BioMedLM + Books + MQG & \textbf{70.56} \\% & \textbf{0.4211}
\bottomrule
\end{tabular}
\caption{Final results for BioMedLM with various parameters. MQG stands for Medical Question Generation. The model is trained on books for three epochs and on MQG for two epochs. All models are then fine-tuned on ECN-QA.}
\label{result}
\end{table}

The results of our experiments are shown in Table~\ref{result}. We report the result of the original BioMedLM, as well as models pre-trained on the collection of books (\textit{BioMedLM + Books}), pre-trained on the questions (\textit{MQG} for Medical Question Generation) and our complete method (\textit{BioMedLM + Books + MQG}). All models are fine-tuned on ECN-QA.

Including books as part of our training data improves the accuracy by approximately 2 points and the MQG method alone by 1 point. The best accuracy is achieved by combining the pre-training using books with the question-generation method. Overall, we significantly improve the baseline with our full method, getting +3 points in accuracy. We also surpass the GPT-3.5 model, as shown in Table~\ref{tab:gpt-results}.

% Furthermore, we intend to explore additional hyperparameters and evaluate the potential of retrieval techniques. Considering the observed challenges in BioMedLM, based on GPT-2 \cite{radford2019language}, which sometimes struggles to focus on the correct parts of questions, we are considering employing newer, larger models like LLaMA 2 \cite{touvronLlamaOpenFoundation2023}. To gain better insights into the model's decision-making process, we also plan to conduct interpretability studies.

\begin{figure}[h]
    \centering
    \includegraphics[width=1.0\linewidth]{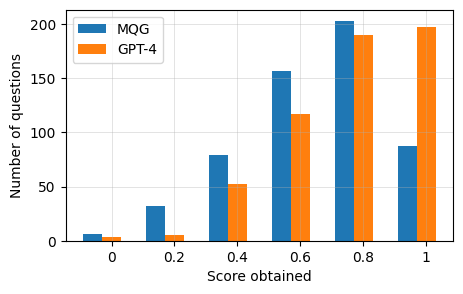}
    \caption{Accuracy distribution by question (number of correct propositions divided by number of total propositions) on the FreeCN dataset of GPT-4 and BioMedLM + Books + MQG
    }
    \label{accdistrib}
\end{figure}

In Figure~\ref{accdistrib}, we display the number of questions for each score for our full method and GPT-4.
We observe that our model still lags behind GPT-4.
Since the model answers all propositions independently and has no knowledge of its answer to other propositions, the model can contradict itself, which makes it harder to obtain a score of 1 (i.e. having the right answer to all propositions). More detailed statistics per subject are available in Figure~\ref{fig:accpersubject}. 
Our method appears less effective on subjects it has not been trained on, such as pediatrics. 

\section{Conclusion}

We introduced ECN-QA, a novel dataset for medical question answering that contains a novel type of exercise: progressive questions. We proposed a training strategy based on prompted question generation that improves results over our baseline model, enabling the model to surpass GPT-3.5 accuracy with a much lower parameter count.

Potential avenues for improving efficient medical question answering include increasing the size of the pre-training dataset and the number of generated questions and investigating retrieval-based answering (open-book exam).
% Additionally, investigating multimodal learning and training directly in low-resource languages instead of translating are possible future directions.
A model with significant capabilities in medical answering can aid in making informed decisions, especially in time-sensitive situations where rapid response is crucial. Such a model can offer up-to-date information, suggest potential diagnoses, and recommend treatment options based on the latest research and clinical guidelines. 

\section{Ethical Concerns}
The model was trained on questions designed for students' examination, not for a real-world clinical setting. The generalization of this model to actual clinical settings is unknown. Indeed the model has potential biases and limitations in handling sensitive and complex medical cases and should not be used as so on real-world patients.

\section{Acknowledgement}

This work was granted access to the HPC resources of IDRIS under the allocation 2023-AD011013489R1 made by GENCI.

\clearpage
\bibliography{main}

\clearpage
\appendix

\section{Related Work}
\label{sec:app:related}

\subsection{Medical QA Datasets}
Various datasets have been developed in medical question answering (QA). Among these, the MedQA dataset \cite{medQA2020} stands out for its comprehensive coverage of multiple-choice questions derived from professional medical board exams. This dataset is particularly significant because it encompasses many questions, totaling 12,723 items. It aims to evaluate the depth of medical knowledge encoded in AI models.

Another dataset is PubMedQA Dataset for Biomedical Questions, \cite{PubMedQADatasetBiomedical2019}. This dataset uniquely focuses on questions generated from article titles and abstracts within the biomedical literature, excluding conclusions, and provides answers in a format conducive to yes/no/maybe evaluations.

Further expanding the landscape, the MedMCQA dataset \cite{MedMCQALargescaleMultiSubject2022} is a large-scale, multi-subject repository of medical multiple-choice questions. This dataset has a large scope and relevance, covering many medical subjects.

\subsection{Medical QA Models}
Several strategies aim to construct a good model with high accuracy and reliability of responses on those medical tests. One method involves leveraging large language models (LLM) such as GPT-4. Through prompt engineering, \cite{brinComparingChatGPTGPT42023} or \cite{noriCapabilitiesGPT4Medical2023a} have demonstrated excellent results on MedQA.

Further exploration into the efficacy of large-scale models has been conducted, with \cite{MedPalm} and \cite{MedPalm2}. These studies have assessed the performance of such models on MedQA and across a diverse array of medical datasets.

Moreover, the landscape of medical QA has been enriched by initiatives to fine-tune pre-existing LLMs. For instance, adaptations of Llama 2~\cite{LLama2} have been proposed~\cite{PMCLLama,Meditron}. These efforts signify a targeted move towards refining the capabilities of LLMs to meet the demands of the medical domain, illustrating a focus on customizing general models for specialized tasks.

In the context of smaller-scale models, \cite{StanfordCRFM} has been recognized for its superior performance. This model stands out as a testament to the effectiveness of more compact models in handling medical QA tasks, offering an alternative to the larger, more complex systems.

% Qwen ?
% Yi ?
% Mistral ?
% Zephyr ?

\section{Question Generation}
\label{sec:app:prompt}

Table~\ref{tab:final-prompt} shows the prompt we used to generate questions with GPT-4. The prompt is appended with a section coming from a medical textbook.

% \noindent
% \framebox{\parbox{\dimexpr\linewidth-2\fboxsep-2\fboxrule}{\itshape%

\begin{table}[h]
\centering
\footnotesize
\begin{tabular}{|m{.95\linewidth}|}
\hline
You are a French professor of medicine. You seek to test the level of medicine of your students. Your task is to generate 1 to 2 different clinical cases requiring the highest medical understanding. Each clinical case consists of an Introduction and 4-10 multiple-choice questions. They must be formatted as follows: Introduction, Propositions. Propositions contain several proposals
 with a justification and a field to know if they are correct.
The clinical case needs to be very very hard and accurate. The level of difficulty is 10 out of 10. It should be very hard even for the best students. And you should have a very detailed justification.
The case should be long with detailed questions and detailed justification.

The criteria to be met are:

1. The introduction is common to all questions.

2. There must be 4-10 different questions.

3. A question can have 5-10 possible choices.

4. One or more proposals may be fair.

5. Justification must be specific, justified and sourced. It is very important to have a very good and long justification. It should be at least 3 lines long.

6. Uses the highest medical level possible.

7. Questions must be diversified to a minimum of 4. They must deal with the patient’s disease but also with the examinations to be carried out, the follow-up and the possible developments of the case. They will make the case both nuanced and complex.

8. The case must be precise or even quantitative. It is a question of providing as much information as possible, and the solution to the questions may be found in detail.

9. Cases must be pedagogical and the questions must be linked to build a complete reasoning.

10. Responses should be directed to prioritize severe and frequent cases.

11. The student’s expected behaviour is above all to avoid medical misconduct.

12. The student’s method must be a probabilistic approach.

13. A language model must be able to answer questions. For example, do not ask the wizard to create images or audio.

14. The case must be written in English.

15. All fields must be completed.

16. The MA for the drug and the recommendations of the HAS and ANSM must be respected. In the absence of recommendations from HAS and ANSM, the current practices recommended by French speciality colleges and learned societies will be applied.

\#\#\#

To do that you can use the following information:
[Extract of a medical book]
\\
\hline
% }}
\end{tabular}
\caption{Prompt used to generate progressive questions with GPT-4}
\label{tab:final-prompt}
\end{table}

\begin{figure*}[t]
    \centering
    \includegraphics[width=1.0\linewidth]{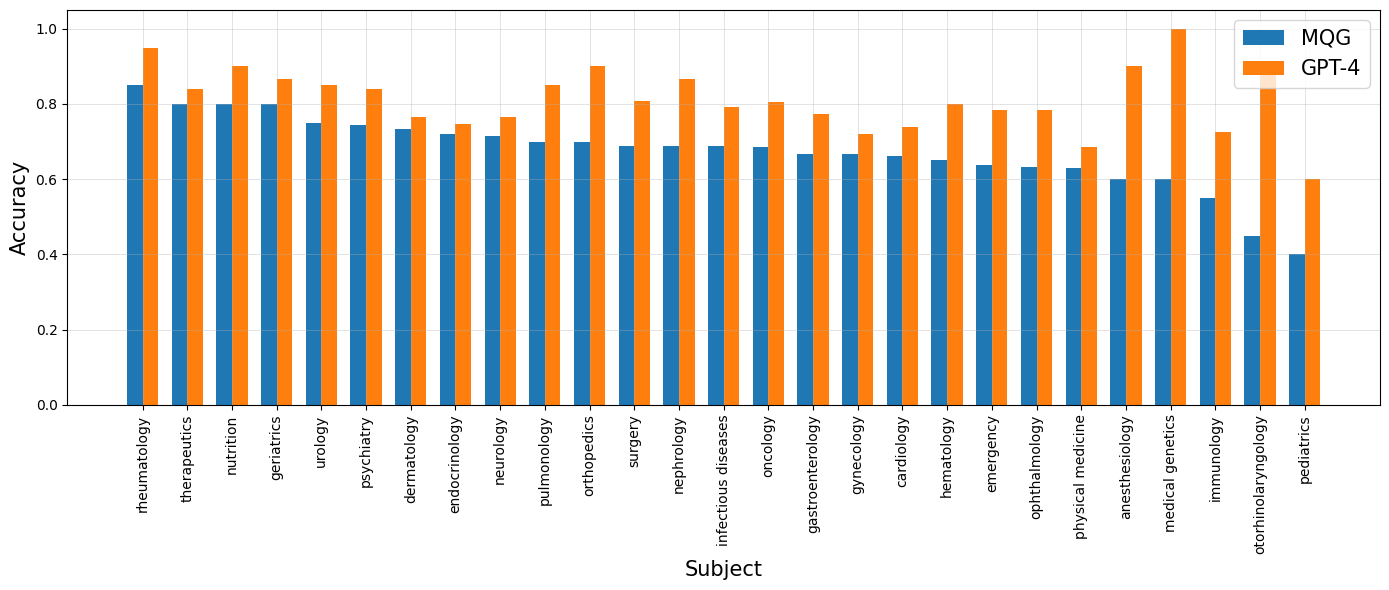}
    \caption{Accuracy per subject of BioMedLM and GPT-4}
    \label{fig:accpersubject}
\end{figure*}

\section{Implementation and training details}
\label{sec:appendix:implementation}
% For training, we leveraged Jean-Zay\footnote{\href{http://www.idris.fr/jean-zay/}{http://www.idris.fr/jean-zay/}}, a supercomputer that provides access to extensive GPU clusters using nodes with 4 V100 of 32 GB. 
For training, we use a node with 4 NVIDIA V100 gpus. The model is pre-trained on books for three epochs on generated cases for two epochs, and we fine-tune the final model for 20 epochs. We use a learning rate of 1.e-4 for pre-training and 2.e-6 for fine-tuning.

% \cor{details about hyperparameters, number of nodes/GPUs, etc}

% \tcbset{colback=white!10!white,colframe=black!5!black,sharp corners}

% \begin{tcolorbox}[breakable,sharp corners,colback=white!10!white,colframe=black!5!black,boxrule=0.1mm]

\section{Additional Results}
\label{sec:supp:additional-results}

We show in Figure~\ref{fig:accpersubject} the accuracies for each topic of the ECN exam for GPT-4 and our model. Our model is close to GPT-4 for most subjects and performs worse on subjects that GPT-4 often refuses to generate questions, like pediatrics. These questions are from the test set but only come from the additional questions provided by the FreeCN team.

\section{Example of Cases}
\label{sec:app:example-case}

In Table~\ref{tab:example-dp}, we display the first question of a progressive question from the ECN-QA dataset, with the propositions of answer. 

\newpage
\begin{table}[t]
\footnotesize
\centering
\begin{tabular}{|m{.95\linewidth}|}
\hline
     \textbf{Introduction}: A 67-year-old man consults for right calf pain occurring after a walk that the patient estimates to be 350 meters away. He is a retired and sedentary taxi driver. This patient has been smoking a pack of cigarettes a day since the age of 30. You follow it for high blood pressure discovered by a systematic and balanced examination by perindopril. Blood sugar is normal as well as lipid balance.
     
\textbf{Question}: What is your main diagnostic hypothesis ? \\
\textbf{Propositions: }
\begin{itemize}
    \item \textbf{Obliterating arterial disease of the lower limbs}
    \item Narrow lumbar canal
    \item Lombosciatica
    \item Hypokalemia
    \item Deep vein thrombosis
\end{itemize}

\textit{For the example the following questions are: }
\begin{itemize}
    \item You suspect arterial disease obliterating the lower limbs. Which of the following semiological elements will guide the diagnosis towards this hypothesis?
    \item The interrogation confirms the appearance of a pain when walking with a cramp localized in the right calf. The pain manifests itself early when the patient climbs a slope, thus supporting your diagnostic hypothesis of arterial obliterating disease of the lower limbs. [...] On the data of this clinical examination, which is(are) the arterial atheromatous lesion(s) that you should suspect?
\end{itemize} \\
\hline
\end{tabular}
\caption{Example of PQ in the ECN-QA dataset. This particular PQ consists of 16 questions, and both this PQ and the previous IQ section are derived from the 2020 ECN. Correct answers are in bold.}
\label{tab:example-dp}
\end{table}

\newtcolorbox{zitat}[2][]{
colback=white,
% grow to right by=-10mm,
% grow to left by=-10mm, 
boxrule=0.5pt,
boxsep=1pt,
breakable,
enhanced jigsaw,
% borderline west={2pt}{0pt}{gray},
colbacktitle=white,
coltitle={black},
fonttitle={\normalsize\bfseries},
sharp corners,
title={#2},
titlerule=0mm,
#1,
}

\clearpage
\subsection{Full Progressive Question}
\label{sec:app:full-dp}
\vspace{-2cm}
Here, we display a full progressive question with all possible answers. Correct answers are in bold font.
\vspace{-5cm}
\begin{zitat}{\vspace{2mm}Full example of a progressive question}
\footnotesize
\textbf{Introduction:} A 54-year-old man, a long-term smoker who has been hypertensive for 12 years (calcium channel blocker treatment), consults his attending physician for an isolated episode of total gross hematuria, without a clot. His other history has been an appendectomy in childhood. The blood count is as follows: Hb 10.4 g/dL (MCV 78 µm3), GB 8 G/L, blisters 247 G/L. Creatinine is 110 µmol/L (estimated glomerular filtration rate of 65 ml/min/1.73 m2). A renal ultrasound showed a hyperechoic mass of 7 cm on the right kidney.

\textbf{Questions:}
What are the elements (present or to be sought at the interrogation and clinical examination) that can evoke a malignant tumor of the kidney? (one or more correct answers)
\begin{itemize}
\item \textbf{Smoking}
\item \textbf{Chronic high blood pressure}
\item Long-term calcium channel blocker treatment
\item A family history of multiple endocrine neoplasia
\item \textbf{Low back pain}
\end{itemize}

Which exam(s) are you asking for as a first line?
\begin{itemize}
\item \textbf{Urinary cytology with pathological examination}
\item \textbf{Cytobacteriological examination of urine}
\item Serum erythropoietin assay
\item \textbf{Abdominopelvic CT scan with and without contrast injection}
\item Ultrasound-guided puncture of the mass
\end{itemize}

On the cut shown below, what are the True propositions?  (one or more correct answers)
\begin{itemize}
\item \textbf{This is an abdominal CT scan with injection}
\item This is a coronal cup
\item Structure number 1 is the inferior vena cava
\item \textbf{The cut passes through the third duodenum}
\item The number 2 corresponds to the inferior mesenteric artery
\end{itemize}

What are the real propositions? (one or more correct answers)
\begin{itemize}
\item \textbf{The patient must receive red blood cells}
\item The patient must receive platelet pellets
\item In case of transfusion of red blood cells, you would prescribe O-negative pellets
\item A search result for irregular agglutinins less than 48 h old must be available
\item Since 2003, there has been no risk of transmission of infectious pathogens through red blood cell transfusion
\end{itemize}

What is the real proposal(s)?
\begin{itemize}
\item \textbf{This is acute renal failure}
\item The glomerular filtration rate must be recalculated
\item An obstacle on the contralateral kidney is likely
\item \textbf{It may be functional renal failure}
\item \textbf{An ionogram should be prescribed on a urine sample}
\end{itemize}

What are the exact proposals? (one or more correct answers)
\begin{itemize}
\item He has moderate chronic renal failure
\item \textbf{His antihypertensive treatment must include an inhibitor of the renin-angiotensin system}
\item The LDL cholesterol target to be achieved is 1.3 g/L
\item He must follow a diet containing no more than 1.5 g/kg of protein weight
\item It is necessary to advocate a diet low in fast sugars
\end{itemize}

What risk(s) does he run?
\begin{itemize}
\item \textbf{Gradual decrease in diuresis}
\item \textbf{Increased cardiovascular risk}
\item \textbf{Hyperphosphoremia}
\item \textbf{Erectile dysfunction}
\item \textbf{Contralateral kidney cancer}
\end{itemize}

What is the True answer(s)?
\begin{itemize}
\item The ALD file is completed by the patient and validated by the medical specialist
\item \textbf{The attending physician must specify in the request the protocol of care envisaged including treatments, examinations and consultations}
\item \textbf{The medical officer of the Health Insurance must validate the care protocol}
\item In case of coverage in ALD, remains the responsibility of the patient only the co-payment
\item The third-party payer is the part of the care paid by the insured whether or not he is registered in ALD
\end{itemize}

What is your interpretation of the electrocardiogram below?
\begin{itemize}
\item \textbf{Sinus rhythm}
\item Sino-auricular block
\item T-waves suggestive of hyperkalemia
\item Expanded QRS Complexes
\item \textbf{Left ventricular hypertrophy}
\end{itemize}

To reduce edematous syndrome, what do you recommend at this stage? (one or more correct answers)
\begin{itemize}
\item A low-salt diet (less than 6 g/d)
\item Water restriction
\item \textbf{A loop diuretic (furosemide)}
\item A thiazide diuretic (hydrochlorothiazide)
\item Blood ultrafiltration (start of hemodialysis)
\end{itemize}

What are the possible cause(s) in the context of the new biological abnormality observed?
\begin{itemize}
\item \textbf{Excessive calcium intake}
\item Taking furosemide
\item \textbf{Chronic renal failure}
\item Secondary hyperparathyroidism
\item \textbf{Bone metastases from kidney cancer}
\end{itemize}

What additional examination(s) do you recommend to explore this biological anomaly?
\begin{itemize}
\item \textbf{Ionized serum calcium}
\item Test de PAK
\item \textbf{PTH assay}
\item PTHrp assay
\item \textbf{Bone scintigraphy}
\end{itemize}

Which proposals are correct? (one or more correct answers)
\begin{itemize}
\item Metastatic cancer is a contraindication to dialysis
\item Haemodialysis confers survival advantage over peritoneal dialysis
\item The preparation of an arteriovenous fistula (AVF) is contraindicated given the prognosis
\item \textbf{A tunneled central venous catheter may be placed to initiate hemodialysis}
\item A transplant from a cadaveric donor must be discussed
\end{itemize}

In general, regarding living donors, what are the real proposals? (one or more correct answers)
\begin{itemize}
\item Only people with a genetic link to the recipient can be donors
\item \textbf{Transplantation can be done in incompatible ABO condition}
\item Rhesus compatibility must be respected
\item HLA incompatibility between donor and recipient is a formal contraindication
\item The donor is remunerated on a basis proportional to the recipient's waiting time
\end{itemize}

\end{zitat}

\section{Extraction of text from PDF files}
\label{sec:app:pdf}

We use the Azure AI Document Intelligence API\footnote{\url{https://azure.microsoft.com/en-us/products/ai-services/ai-document-intelligence}} to extract text sections from PDF files. The API returns paragraphs of texts and titles, sorted in reading order, along with tables and figures. We remove all tables and figures and implement text filtering algorithms to remove useless paragraphs, like headers, footers, or paragraphs that are just a few characters long.

We then regroup paragraphs in sections based on the titles and divide them into subsections, or regroup them, to have sections of similar lengths, between 500 and 1000 words.

\section{Generated Progressive Question}
\label{sec:app:generated-dp}
Below, we present an example of a progressive question generated by GPT-4.  Correct answers are in bold font.

\begin{zitat}{}
\footnotesize
\textbf{Introduction:} A 45-year-old female is being admitted to the emergency department. She is complaining of a severe and sudden headache unlike any she has ever experienced before. The headache was followed by episodes of vomiting and photophobia. Her Glasgow Coma Scale (GCS) score on admission is 14 and her physical examination is unremarkable. Computed Tomography (CT) of the brain reveals subarachnoid hemorrhage (SAH). \\
\textbf{Questions}
What is the most likely diagnosis in this patient?
\begin{itemize}
    \item \textbf{Ruptured cerebral aneurysm} (The presence of SAH in the setting of a severe, sudden-onset headache (often described as a 'thunderclap headache'), especially when associated with vomiting and photophobia, is highly suggestive of a ruptured cerebral aneurysm. Other common symptoms of SAH due to a ruptured aneurysm include neck stiffness, altered mental status and cranial nerve palsies.)
    \item Migraine (While migraines can cause severe headaches, they do not typically lead to SAH. Moreover, the acute onset described in the vignette is uncharacteristic of migraines which usually have a more gradual onset.)
    \item Tension headache (A tension headache often presents as a dull, constant and bilateral headache. SAH or sudden severe headache is not typical of a tension headache.)
    \item Cluster headache (Cluster headaches are usually unilateral, severe, and around the eye, and are associated with autonomic symptoms such as tearing, nasal congestion and ptosis. They do not typically cause vomiting or lead to SAH.)
    \item Brain tumor (While brain tumors can cause headaches, they do not typically present acutely. Moreover, they generally result in neurological symptoms and signs that localize to the area of the brain where the tumor is located rather than leading to SAH.)
\end{itemize}
What is the next step in the management of this patient?
\begin{itemize}
    \item \textbf{Cerebral angiography} (Cerebral angiography is the gold standard for diagnosing cerebral aneurysms. It can demonstrate the size, location and morphology of the aneurysm and help plan intervention. Cerebral angiography is also able to diagnose other causes of SAH such as arteriovenous malformation.)
    \item Magnetic resonance imaging (MRI) (While MRI may be used to evaluate for other causes of SAH such as arteriovenous malformations (AVM) if the angiogram is inconclusive cerebral angiography is the preferred primary imaging modality in this clinical scenario.)
    \item Lumbar puncture (Lumbar puncture may be performed when the clinical suspicion for SAH is high but initial CT scan is negative. Given the positive CT findings in this case lumbar puncture would not be the best next steps.)
    \item Blood cultures (Blood cultures are indicated in patient with suspected infective endocarditis that can lead to mycotic aneurysms. In this case there is no clinical suspicion of infective endocarditis.)

    \item Repeat CT scan (A repeat CT scan may be used in the surveillance of SAH treatment and complications but it is not the best choice for the acute management of suspected ruptured cerebral aneurysm which requires immediate intervention.)
\end{itemize}     
If this patient’s cerebral angiography reveals a vasospasm of the cerebral arteries, what is the most appropriate step that should be taken?
\begin{itemize}
    \item Platelet antiaggregation (Platelet antiaggregation is not the standard management for cerebral vasospasm. Studies showed that antiplatelet medication would lead to a higher rate of rebleeding in cerebral aneurysm.)
    \item Anticoagulation (Anticoagulation is not the standard management for cerebral vasospasm as it could increase the risk of bleeding.)
    \item \textbf{Calcium channel blockers} (Nimodipine, a calcium channel blocker, is the only medication proven to improve neurologic outcomes in the case of SAH. It doesn’t prevent or treat the vasospasm itself but reduces the severity of neurologic deficits and improves outcome.)
    \item Statins (Although statins have been shown to have some vasoprotective effects they are not the first-line treatment for cerebral vasospasm.)
    \item Analgesics (While managing pain is an important component of care in this patient population analgesics themselves do not treat vasospasm).
\end{itemize}
\end{zitat}

\end{document}